\pdfoutput=1

\documentclass[11pt]{article}

\usepackage[]{acl}

\usepackage{times}
\usepackage{latexsym}

\usepackage[T1]{fontenc}

\usepackage[utf8]{inputenc}

\usepackage{microtype}
\usepackage{subfigure}

\usepackage{array}
\usepackage{multirow}
\usepackage{booktabs}
\usepackage{graphicx}
\usepackage{amssymb}
\usepackage{algorithm}
\usepackage{times}
\usepackage{latexsym}

\usepackage{helvet} 
\usepackage{courier}  
\usepackage{graphicx} 
\usepackage{natbib}  
\usepackage{caption} 

\usepackage{graphicx}
\usepackage{booktabs}
\usepackage{subfigure}
\usepackage{algorithm}
\usepackage{algorithmic}
\usepackage{color}
\renewcommand{\algorithmicrequire}{\textbf{Input:}}

\usepackage{microtype}
\usepackage{amsmath, amssymb, amscd, amsfonts}
\usepackage{IEEEtrantools}
\usepackage{soul}
\usepackage{url}
\usepackage{algorithm}
\usepackage{algorithmic}
\usepackage{amssymb}
\usepackage{amsfonts} 
\usepackage{multirow}
\usepackage{boldline}
\usepackage{hhline}
\usepackage{xcolor}

\usepackage{makecell}
\usepackage{mathrsfs}
\usepackage[normalem]{ulem}
\useunder{\uline}{\ul}{}
\usepackage[inline]{enumitem}
\usepackage{enumitem}

\usepackage[T1]{fontenc}

\usepackage[utf8]{inputenc}

\usepackage{microtype}
\title{An Information Minimization Based Contrastive Learning Model for Unsupervised Sentence Embeddings Learning}

\author{Shaobin Chen$^{1}$, Jie Zhou$^{2}$, Yuling Sun$^{1}$\thanks{\quad Yuling Sun is the corresponding authors of this paper.} \and Liang He$^1$ \\
        $^1$School of Computer Science and Technology, East China Normal University Shanghai, China \\
        $^2$School of Computer Science, Fudan Univerisity Shanghai, China \\ 
        shaobin\_chen@stu.ecnu.edu.cn,
        jie\_zhou@fudan.edu.cn,
        \{ylsun, lhe\}@cs.ecnu.edu.cn}
        
\begin{document}
\maketitle
\begin{abstract}
Unsupervised sentence embeddings learning has been recently dominated by contrastive learning methods (e.g., SimCSE), which keep positive pairs similar and push negative pairs apart. 
The contrast operation aims to keep as much information as possible by maximizing the mutual
information between positive instances, which leads to redundant information in sentence embedding. To address this problem, we present an information minimization based contrastive learning (\texttt{InforMin-CL}) model to retain the useful information and discard the redundant information by maximizing the mutual
information and minimizing the information entropy between positive instances meanwhile for unsupervised sentence representation learning. 
Specifically, we find that information minimization can be achieved by simple contrast and reconstruction objectives.
The reconstruction operation reconstitutes the positive instance via the other positive instance to minimize the information entropy between positive instances. We evaluate our model on fourteen downstream tasks, including both supervised and unsupervised (semantic textual similarity) tasks. Extensive experimental results show that our \texttt{InforMin-CL} obtains a state-of-the-art performance. Code is made available. \footnote{https://github.com/Bin199/InforMin-CL}
\end{abstract}

\section{Introduction}

How to learn universal sentence embeddings by large-scale pre-trained models \cite{devlin2019bert,liu2019roberta}, such as BERT, has been studied extensively in the literature \cite{Gao:Embeddings,reimers2019sentence}. Recently, contrastive learning has been used widely to learn better sentence embeddings \cite{meng2021cocolm,Gao:Embeddings}. Generally, contrastive learning uses various data augmentation methods to generate different views of the input sentences and samples positive instances and negative instances from views. Contrastive learning aims to learn effective embeddings by pulling positive instances together and pushing positive and negative instances apart. This operation focus on maximizing the mutual information between positive instances to retain as much information as possible, which includes both the useful and useless information. Previous studies like \cite{meng2021cocolm,Gao:Embeddings} ignore the redundant information stored in views, which has a bad impact on the performance of downstream tasks, as proved by \cite{achille2018emergence,conf/nips/Tian0PKSI20}.

Table \ref{Instance} gives an example of redundant information stored in training texts. In training texts, the sentences contain much redundant information which is not favorable to the downstream task. The redundant information may be stop words and the style of the sentence (e.g., restatement, capitalization, and hyphen).
The existing study \cite{conf/nips/Tian0PKSI20} also shows that discarding redundant information in views can help to improve the performance of the downstream task.
Thus, we arise the following question: how to discard this redundant information by choosing the optimal views?

\begin{table}[!t]
\small
    \caption{Training texts may contain various kinds of redundant information, such as stop words, restatement, capitalization, and hyphen.}
    \label{Instance}
    \centering
    \begin{tabular}{ll}
    \toprule[2pt]
    Original & Where is the party, it sounds great.\\
    \midrule[1pt]
    Stop words & Where \colorbox{orange}{is the} party, \colorbox{orange}{it} sounds great.\\
    Restatement & The party sounds great, where is it.\\
    Capitalization & Where Is The Party, It Sounds Great.\\
    Hyphen & Where-is-the-party, it-sounds-great.\\
    \bottomrule[2pt]
    \end{tabular}
\end{table}

It is natural to solve the above questing via information bottleneck (IB), which has been utilized as an effective and simple method for learning a good embedding by keeping the important information and forgetting redundant information in various tasks \cite{tishby1999information,chen2020learning,tishby2015deep}. 
Prompted by this, we attend to solve the problem by drawing inspiration from the information minimization principle (an idea in IB theory) \cite{conf/nips/Tian0PKSI20}: A good set of views share the minimal information necessary to perform well at the downstream task. 
This method aims to retain useful information and forget redundant information.
In this paper, we explore how to address the shortcomings (i.e., ignoring the redundant information stored in views) of previous work for unsupervised sentence representations via the information minimization principle.

We propose an information minimization based contrastive learning (\texttt{InforMin-CL}) model for unsupervised sentence embedding learning. 
\texttt{InforMin-CL} incorporates the information minimization principle into contrastive learning to not only learn the important information but also drop the redundant information.
We optimize our \texttt{InforMin-CL} model from two perspectives: contrast and reconstruction.
Firstly, we learn the useful information by a contrast task to maximize the mutual information. This task manages to attract positive pairs and repulse negative pairs. Attracting positive pairs stands for maximizing mutual information between positive instances. Secondly, we propose a reconstruction task that encourages the model to reconstruct the representation of the positive instance via the other one in the same pair. 


The algorithm of \texttt{InforMin-CL} is easy to understand and can be implemented with just several lines of code. Moreover, our method does not change the major network structure, so it is model-agnostic and can be applied to any representation learning neural networks based on contrastive learning. Experiments in Section \ref{section: experiment} show that \texttt{InforMin-CL} can help the model learn effective representations that improve downstream task performance. Our main contributions are summarized as follows: \begin{itemize}
    \item We propose an \texttt{InforMin-CL} model to learn good sentence embeddings by keeping useful information and getting rid of redundant information between positive instances.
    \item We achieve our model via simple contrast and reconstruction tasks and prove that the reconstruction task can drop redundant information by minimizing the information entropy. 
    \item We achieve new state-of-the-art results on seven supervised tasks and seven unsupervised tasks, which indicates the great advantages of our proposed model.
\end{itemize}

\begin{figure*}[!t]
    \centering
    \includegraphics[width=\textwidth]{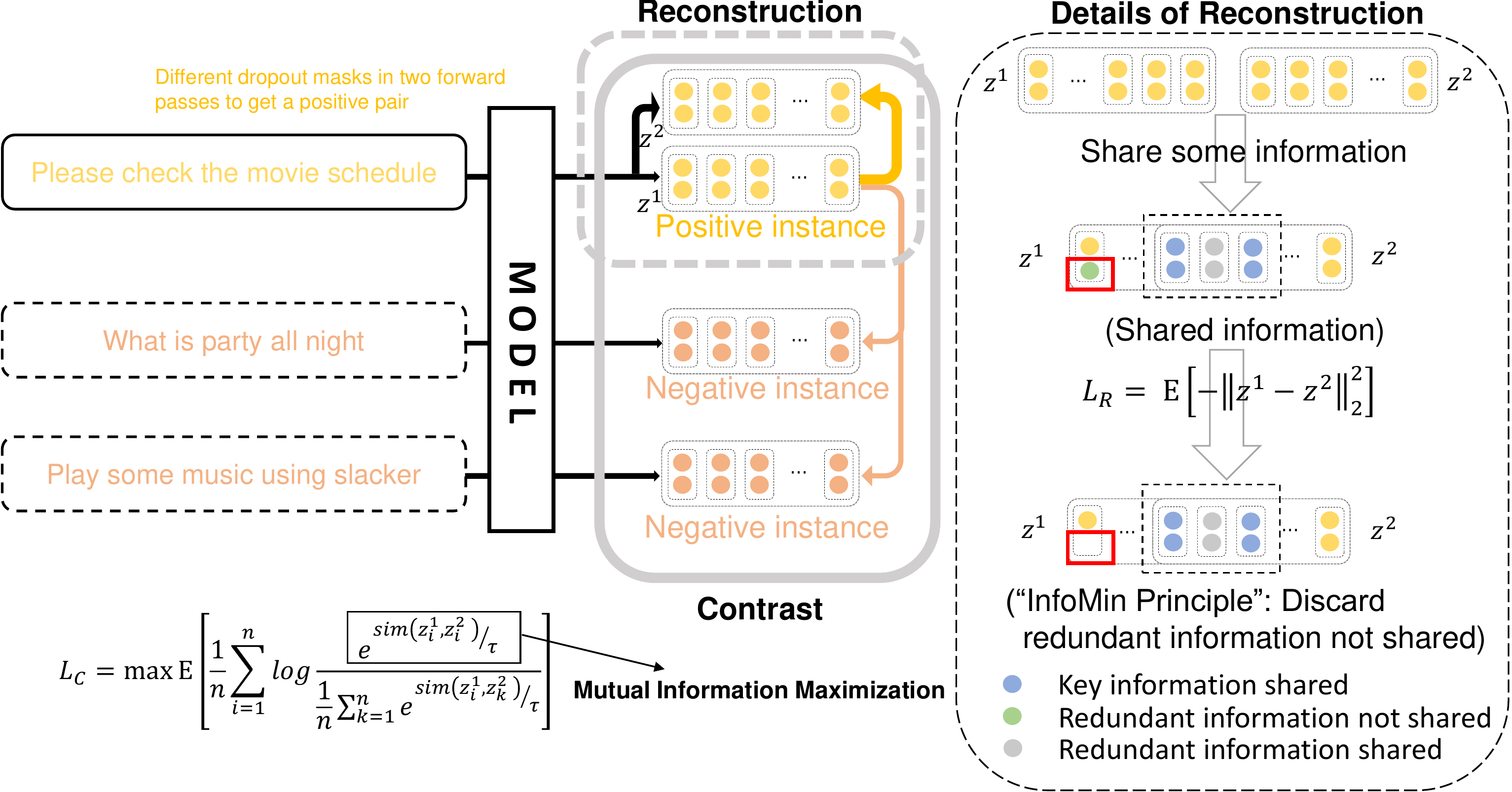}
    \caption{The architecture of our proposed framework. Contrast: Negative pairs are pushed apart while positive pairs are pulled together, which suggests maximizing the mutual information between positive instances. Reconstruction: We drop redundant information in $z^1$ not shared with $z^2$ (marked by the green point) by reconstructing one positive instance via the other positive instance.}
    \label{Fig1}
\end{figure*}

\section{Related Work}

\subsection{Sentence Representation Learning}
Learning sentence embeddings as an important problem in NLP has been widely studied. Recent work focus on leveraging the power of BERT \cite{devlin2019bert} to learn effective sentence embeddings, which are free from artificially supervised signals. 
BERT-flow \cite{conf/emnlp/LiZHWYL20} transforms the anisotropic sentence embedding distribution into a smooth isotropic Gaussian distribution through normalizing flows. BERT-whitening \cite{su2021whitening} further presents a whitening operation to enhance the isotropy of sentence embeddings and achieves better results. 

Then, the contrastive learning approach is applied for sentence embedding learning.
IS-BERT \cite{zhang2020unsupervised} proposes a model with a feature extractor on top of BERT and an objective that maximizes the mutual information between global sentence embeddings and local sentence embeddings. CLEAR \cite{wu2020clear} employs multiple sentence-level augmentation strategies to learn sentence representation. Coco-LM \cite{meng2021cocolm} employs an auxiliary language model to corrupt text sequences, upon which it constructs a token-level task and a sequence-level task for pre-training the main model. Gao \textit{et al.}~\cite{Gao:Embeddings} presents an unsupervised approach that predicts input itself with dropout noise and a supervised approach utilizing natural language inference datasets. SCD \cite{klein2022scd} leverages the self-contrast of augmented samples obtained by dropout, which eliminates the reliance on negative pairs. However, all these studies lack consideration of discarding redundant information stored in views. In our work, we consider and solve the problem in the framework of the information minimization principle. 

\subsection{Information Minimization Principle}
Information minimization principle \cite{conf/nips/Tian0PKSI20} has been proposed to retain the minimal information necessary. In recent years, researchers utilize the information minimization principle to improve image representations \cite{conf/nips/Tian0PKSI20,conf/iclr/Tsai0SM21}. 
Furthermore, information bottleneck is used to improve the interpretability of the attention-based models \cite{zhou2021attending}.
Tian \textit{et al.} \cite{conf/nips/Tian0PKSI20} shows good views for a given task in a contrastive representation learning framework should retain task-relevant information while minimizing irrelevant nuisances. However, it focuses on eliminating the task-irrelevant information via downstream datasets. Tsai \textit{et al.} \cite{conf/iclr/Tsai0SM21} focuses on the multi-view setting between input and self-supervised signals and adopts self-supervised signals to reconstruct learned representations to discard task-irrelevant information. Our work also differs from \cite{conf/iclr/Tsai0SM21} in two perspectives: 1) we discard redundant information in texts while \cite{conf/iclr/Tsai0SM21} drops noise information stored in images, and additionally different self-supervised signals are used in two work; 2) \cite{conf/iclr/Tsai0SM21} validates their method by learning visual features evaluated by supervised tasks while sentence embedding learning evaluates embeddings via not only supervised tasks but also unsupervised tasks. 

\section{Method}
We propose an \texttt{InforMin-CL} model for unsupervised sentence representation learning (Figure \ref{Fig1}). There are two main steps, contrast and reconstruction. We first present the contrast objective to learn the useful information: we push apart positive instances and negative instances while pulling positive instances together, which implies maximizing mutual information between positive instances. Later we present the reconstruction task to drop the useless information: we minimize the conditional information entropy of one positive instance given the other positive instance. Algorithm \ref{algorithm 1} provides the pseudo-code of \textit{InforMin-CL}.

\renewcommand\algorithmicrequire{\textbf{Input:}}
\begin{algorithm}
    \small
    \caption{Pseudocode of \textit{InforMin-CL} in a PyTorch-like style.}
    \label{algorithm 1}
    \begin{algorithmic}
        \REQUIRE{batch size N, temperature $\tau$, structure of $f$ and $\Gamma $.}
        \ENSURE{encoder network $f\left(  \cdot  \right)$.}
        \FOR{sampled minibatch $\left\{ {{x_k}} \right\}_{k = 1}^N$}
            \FORALL{$k \in \left\{ {1, \ldots ,N} \right\}$}
                \STATE draw two augmentation functions $t \sim \Gamma $, $t' \sim \Gamma $ 
                \STATE ${\tilde x_{2k - 1}} = t\left( {{x_k}} \right)$
                \STATE ${z_{2k - 1}} = f\left( {{{\tilde x}_{2k - 1}}} \right)$
                \STATE ${\tilde x_{2k}} = t'\left( {{x_k}} \right)$
                \STATE ${z_{2k}} = f\left( {{{\tilde x}_{2k}}} \right)$
            \ENDFOR
            \FORALL {$i \in \left\{ {1, \ldots ,2N} \right\}$ and $j \in \left\{ {1, \ldots ,2N} \right\}$}
                \STATE ${s_{i,j}} = z_i^{\rm T}{z_j}/\left( {\left\| {{z_i}} \right\|\left\| {{z_j}} \right\|} \right)$
            \ENDFOR
            \STATE ${L_C} = \max {\rm{E}}\left[ {\frac{1}{N}\sum\limits_{i = 1}^N {\log \frac{{\exp \left( {{{{s_{2i - 1,2i}}} \mathord{\left/
 {\vphantom {{{s_{2i - 1,2i}}} \tau }} \right.
 \kern-\nulldelimiterspace} \tau }} \right)}}{{\frac{1}{N}\sum\limits_{k = 1}^N {\exp \left( {{{{s_{2i - 1,2k}}} \mathord{\left/
 {\vphantom {{{s_{2i - 1,2k}}} \tau }} \right.
 \kern-\nulldelimiterspace} \tau }} \right)} }}} } \right]$
            \STATE ${L_R} = {\rm E}\left[ { - \left\| {{z_{2k - 1}} - {z_{2k}}} \right\|_2^2} \right]$
            \STATE $L = {L_C} + {L_R}$
            \STATE update $f$ to minimize $L$
        \ENDFOR
    \end{algorithmic}
\end{algorithm}

\subsection{Contrast}
Contrastive learning attends to learning effective representation by pulling positive sample pairs together and pushing apart negative sample pairs. 
We build upon the recent success of unsupervised SimCSE \cite{Gao:Embeddings} and take the embeddings derived from the same sentence with independently different dropout masks as positive instances. 
We adopt the dropout mask (with default dropout probability p = 0.1) as an augmentation skill, which is proved to outperform other skills \cite{Gao:Embeddings}, to obtain positive pairs. 
The positive pair takes the same sentence, and their embeddings only differ in dropout masks. Other sentences in the same mini-batch are seen as negative instances. 

We denote input, instance, and self-supervised signal as X, Z, and S. We feed the same input x to the encoder twice by applying different dropout masks and then get positive instances ${z^1}$ and ${z^2}$. In our work, we take one instance $z^2$ in positive pair as a self-supervised signal. The information required for downstream tasks is referred to as ``key information": $T$. $I$ and $H$ represent mutual information and information entropy. 

Let $Z^{sup}$ be the sufficient supervised representation and $Z^{su{p_{\min }}}$ be the minimal and sufficient supervised representation:
\begin{equation}
\begin{aligned}
& {Z^{\sup }} = \mathop {\arg \max }\limits_Z I\left( {Z;T} \right) \\
& {Z^{su{p_{\min }}}} = \mathop {\arg \min }\limits_Z H\left( {Z|T} \right) \\
& s.t.\,\,\,I\left( {Z;T} \right)is\,{\rm{maximized}}
\end{aligned}
\end{equation}

Let $Z^{ssl}$ be the sufficient self-supervised representation and $Z^{ss{l_{\min }}}$ be the minimal and sufficient self-supervised representation:
\begin{equation}
\begin{aligned}
& {Z^{ssl}} = \mathop {\arg \max }\limits_Z I\left( {Z;S} \right) \\
& {Z^{ss{l_{\min }}}} = \mathop {\arg \min }\limits_Z H\left( {Z|S} \right) \\
& s.t.\,\,\,I\left( {Z;S} \right)is\,{\rm{maximized}}
\end{aligned}
\label{Eq define_ssl}
\end{equation}

Then, we give theorem 1. For proof, please refer to \cite{conf/iclr/Tsai0SM21}.\\
\textbf{Theorem 1} \textit{The supervised learned representations
contain all the key information in the input (i.e. $I\left( {X;T} \right)$). The self-supervised learned representations 
contain all the key information in the input with a potential loss ${\varepsilon}$:}
\begin{equation}
\begin{aligned}
I\left( {X;T} \right)  &  = I\left( {{Z^{\sup }};T} \right) = I\left( {{Z^{{{\sup }_{\min }}}};T} \right)\\
 & \ge I\left( {{Z^{ssl}};T} \right) \\
 & \ge I\left( {{Z^{ss{l_{\min }}}};T} \right) \\
 & \ge I\left( {X;T} \right) - \varepsilon 
\end{aligned}
\end{equation}



The contrastive learning objective maximizes the dependency between positive instance ${z^1}$ and self-supervised signal ${z^2}$, which suggests maximizing the mutual information $I\left( {{z^1};{z^2}} \right)$. Theorem 1 suggests that maximizing $I\left( {{z^1};{z^2}} \right)$ results in ${z_1}$ containing almost all the information required for downstream tasks from the input x. Note that T is utilized only for describing our method and in practice, no downstream datasets are used in the pre-training phase. 

We use a contrastive learning objective similar to that in \cite{oord2018representation}, which is a mutual information lower bound with low variance:
\begin{equation}
    {{\mathcal L}_{C}} = \max {\mkern 1mu} {\mathbb{E}}\left[ {\frac{1}{N}\sum\limits_{i = 1}^N {\log \frac{{{\rm{ }}{\raise0.5ex\hbox{$\scriptstyle {{e^{sim\left( {z_i^1,z_i^2} \right)}}}$}
\kern-0.1em/\kern-0.15em
\lower0.25ex\hbox{$\scriptstyle \tau $}}}}{{\frac{1}{N}\sum\nolimits_{k = 1}^N {{\raise0.5ex\hbox{$\scriptstyle {{e^{sim\left( {z_i^1,z_k^2} \right)}}}$}
\kern-0.1em/\kern-0.15em
\lower0.25ex\hbox{$\scriptstyle \tau $}}{\rm{ }}} }}} } \right]
    \label{Eq 5}
\end{equation}
where $\left( {z_1^1,z_1^2} \right), \ldots ,\left( {z_N^1,z_N^2} \right) \sim {P^N}\left( {{Z^1},{Z^2}} \right)$, $z_i^1, z_i^2$ are two positive instances of the $i$-th examples. N refers to batch size and P refers to the statistical distribution of $\left( {{Z^1},{Z^2}} \right)$.

\subsection{Reconstruction}
\label{sect:reconstruction}
The details of the reconstruction are illustrated in the right of Figure \ref{Fig1}. 
The positive instances $z^1$ and $z^2$ contain independent information while sharing some information. The noise information (marked as the green point) in $z^1$ is expected to be discarded by the reconstruction task. We prove that this task can discard the useless information by minimizing the information entropy.

The reconstruction task encourages the self-supervised signal $z^2$ to reconstruct the learned representation ${z^1}$, which suggests maximizing the log conditional likelihood ${{\mathbb{E}}_{{P_{Z^1,{Z^2}}}}}\left[ {\log P\left( {{Z^1}|Z^2} \right)} \right]$. We know that
\begin{equation}
    - H\left( {{Z^1}|Z^2} \right) = {{\mathbb{E}}_{{P_{Z^1,{Z^2}}}}}\left[ {\log P\left( {{Z^1}|Z^2} \right)} \right]
\end{equation}

Thus, this reconstruction also means minimizing $H\left( {{Z^1}|{Z^2}} \right)$.\\
\textbf{Theorem 2} \textit{The sufficient self-supervised representation
contains more redundant information in the input than the sufficient and minimal self-supervised representation.
The latter contains an amount of the information, $I\left( {X;S|T} \right)$, that cannot be discarded from the input:}
\begin{equation}
\begin{aligned}
 I\left( {{Z^{ssl}};X|T} \right) & = I\left( {X;S|T} \right) + I\left( {{Z^{ssl}};X|S,T} \right)\\
  & \ge I\left( {{Z^{ss{l_{\min }}}};X|T} \right) = I\left( {X;S|T} \right)\\
 & \ge I\left( {{Z^{{{\sup }_{\min }}}};X|T} \right) = 0
\end{aligned}
\end{equation}

Theorem 2 (please refer to \cite{conf/iclr/Tsai0SM21} for proof) indicates that ${Z^{ssl}}$ contains two parts of redundant information while ${Z^{ss{l_{\min }}}}$ contains one part of redundant information, discarding $I\left( {{Z^{ssl}};X|S,T} \right)$.

\begin{table*}[h]
\small
\centering
\caption{Unsupervised task results (spearman's correlation). $\dag$: results from \protect\cite{Gao:Embeddings}. $\heartsuit$: results from \protect\cite{klein2022scd}. All other results are reproduced and reevaluated by ourselves.} 

\label{Unsupervised downstream task results}

\setlength{\tabcolsep}{1.3mm}{\begin{tabular}{lcccccccc}
\toprule[2pt]
\textbf{Model} & \textbf{STS12} & \textbf{STS13} & \textbf{STS14} & \textbf{STS15} & \textbf{STS16} & \textbf{STS-B} & \textbf{SICK-R} & \textbf{Avg.} \\
\midrule[1pt]
${{\rm{GloVe\,embeddings }}\left( {{\rm{avg}}{\rm{.}}} \right)}^{\dag}$ & 55.14 & 70.66 & 59.73 & 68.25 & 63.66 & 58.02 & 53.76 & 61.32 \\
${{{\rm{BER}}{{\rm{T}}_{{\rm{base}}}}\left( {{\rm{first - last}}\left. {{\rm{avg}}{\rm{.}}} \right)} \right.}^{\dag}}$ & 39.70 & 59.38 & 49.67 & 66.03 & 66.19 & 53.87 & 62.06 & 56.70\\
${{\rm{BER}}{{\rm{T}}_{{\rm{base}}}}{\rm{ - flow}}^{\dag}}$  & 58.40 & 67.10 & 60.85 & 75.16 & 71.22 & 68.66 & 64.47 & 66.55\\
${{\rm{BER}}{{\rm{T}}_{{\rm{base}}}}{\rm{ - whitening}}^{\dag}}$ & 57.83 & 66.90 & 60.90 & 75.08 & 71.31 & 68.24 & 63.73 & 66.28 \\
${{\rm{IS - BER}}{{\rm{T}}_{{\rm{base}}}}^{\dag}}$ & 56.77 & 69.24 & 61.21 & 75.23 & 70.16 & 69.21 & 64.25 & 66.58 \\ 
${{\rm{CT - BER}}{{\rm{T}}_{{\rm{base}}}}^{\dag}}$ & 61.63 & 76.80 & 68.47 & 77.50 & 76.48 & 74.31 & 69.19 & 72.05\\
${\rm{SCD - BER}}{{\rm{T}}_{{\rm{base}}}}^{\heartsuit}$ & 66.94 & 78.03 & 69.89 & 78.73 & 76.23 & 76.30 & \textbf{73.18} & 74.19\\
${{\rm{SimCSE - BER}}{{\rm{T}}_{{\rm{base}}}}}$ & 67.01 & 82.14 & 73.76 & 80.49 & 79.01 & 77.04 & 69.94 & 75.63 \\
${\rm{\texttt{InforMin-CL} - BER}}{{\rm{T}}_{{\rm{base}}}}$ & \textbf{70.22} & \textbf{83.48} & \textbf{75.51} & \textbf{81.72} & \textbf{79.88} & \textbf{79.27} & 71.03 & \textbf{77.30} \\
\midrule[1pt]
${{{\rm{RoBERT}}{{\rm{a}}_{{\rm{base}}}}\left( {{\rm{first - last\, avg}}{\rm{.}}} \right)}^{\dag}}$ & 40.88 & 58.74 & 49.07 & 65.63 & 61.48 & 58.55 & 61.63 & 56.57\\
${{\rm{RoBERT}}{{\rm{a}}_{{\rm{base}}}}{\rm{ - whitening}}^{\dag}}$ & 46.99 & 63.24 & 57.23 & 71.36 & 68.99 & 61.36 & 62.91 & 61.73\\
${{\rm{DeCLUTR - RoBERT}}{{\rm{a}}_{{\rm{base}}}}^{\dag}}$ & 52.41 & 75.19 & 65.52 & 77.12 & 78.63 & 72.41 & 68.62 & 69.99\\
${\rm{SCD - RoBERT}}{{\rm{a}}_{{\rm{base}}}}^{\heartsuit}$ & 63.53 & 77.79 & 69.79 & 80.21 & 77.29 & 76.55 & 72.10 & 73.89\\
${\rm{SimCSE - RoBERT}}{{\rm{a}}_{{\rm{base}}}}$ & \textbf{70.32} & 82.48 & \textbf{74.84} & \textbf{82.13} & \textbf{82.14} & \textbf{81.57} & 68.62 & \textbf{77.44} \\
${\rm{\texttt{InforMin-CL} - RoBERT}}{{\rm{a}}_{{\rm{base}}}}$ & 69.79 & \textbf{82.57} & 73.36 & 80.91 & 81.28 & 81.07 & \textbf{70.30} & 77.04 \\
\midrule[1pt]
${\rm{SimCSE - RoBERT}}{{\rm{a}}_{{\rm{large}}}}{\mkern 1mu}$ & \textbf{72.64} & 83.78 & \textbf{75.83} & \textbf{84.24} & \textbf{80.12} & \textbf{81.10} & 69.81 & \textbf{78.22} \\
${\rm{\texttt{InforMin-CL} - RoBERT}}{{\rm{a}}_{{\rm{large}}}}$ & 70.91 & \textbf{84.20} & 75.57 & 82.26 & 79.68 & \textbf{81.10} & \textbf{72.81} & 78.08 \\
\bottomrule[2pt]
\end{tabular}}
\end{table*}

Thus, if $z^2$ can perfectly reconstruct ${z^1}$ for any
\begin{equation}
    \left( {z^1,{z^2}} \right) \sim {P_{Z^1,{Z^2}}}
\end{equation}
under the constraint that $I\left( {{z^1};z^2} \right)$ is maximized, we get ${z^{{1_{ss{l_{\min }}}}}}$ according to Eq. \ref{Eq define_ssl}. And then ${z^1}$ discards redundant information, excluding $I\left( {z^1;z^2|t} \right)$ (i.e., the amount of redundant information in the shared information between two positive instances $z^1$ and $z^2$). For easier optimization, we use ${{{\rm \mathbb{E}}}_{{P_{{Z^1},{Z^2}}}}}\left[ {\log {Q_\Phi }\left( {{Z^1}|Z^2} \right)} \right]$ as the lower bound of ${{\mathbb{E}}_{{P_{Z^1,{Z^2}}}}}\left[ {\log P\left( {{Z^1}|Z^2} \right)} \right]$. In our deployment, we utilize the design in Eq. \ref{Eq 5} and let ${{Q_\Phi }\left( {{Z^1}|Z^2} \right)}$ be Gaussian $N\left( {{Z^1}|Z^2,\sigma {\bf{{\rm I}}}} \right)$ with $\sigma {\bf{{\rm I}}}$ as a diagonal matrix. Hence, we obtain the reconstruction objective as follows: 
\begin{equation}
    {\mathcal{L}_{R}} = {{\rm{\mathbb{E}}}_{{z^1},{z^2} \sim \;{P_{{Z^{1,}}{Z^2}}}}}\left[ { - \left\| {{z^1} - {z^2}} \right\|_2^2} \right]
\end{equation}

We combine two objectives as a total objective: 
\begin{equation}
    {\mathcal{L}} = {\mathcal{L}_{C}} + \lambda *{\mathcal{L}_{R}}
    \label{Eq loss}
\end{equation}
where $\lambda $ is a hyper-parameter. Training model with the total loss enables us to discard redundant information in views. 

\begin{table}
    \centering
        \caption{Batch sizes, learning rates and $\lambda$ adopted for \texttt{InforMin-CL}.}
    \label{batch sizes, learning rates and lambda}
    \begin{tabular}{cccc}
    \toprule[2pt]
    {} & {BERT} & \multicolumn{2}{c}{RoBERTa} \\
    \midrule[1pt]
    {} & base & base & large\\
    \midrule[1pt]
    Batch size & 128 & 128 & 128\\
    Learning rate & 3e-5 & 1e-5 & 3e-5 \\
    $\lambda$ & 0.4 & 4 & 4\\
    \bottomrule[2pt]
    \end{tabular}
\end{table}

\begin{table*}[h]
\small
\centering
\caption{Supervised task results (accuracy). $\dag$: results from \protect\cite{Gao:Embeddings}. $\heartsuit$: results from \protect\cite{klein2022scd}. All other results are reproduced and reevaluated by ourselves. w/ MLM: adding MLM as an auxiliary task with $\beta = 0.1$.}
\label{Supervised downstream task results}

\setlength{\tabcolsep}{1.5mm}{\begin{tabular}{lcccccccc}
\toprule[2pt]
\textbf{Model} & \textbf{MR} & \textbf{CR} & \textbf{SUBJ} & \textbf{MPQA} & \textbf{SST} & \textbf{TREC} & \textbf{MRPC} & \textbf{Avg.} \\
\midrule[1pt]
${{\rm{GloVe\,embeddings }}\left( {{\rm{avg}}{\rm{.}}} \right)}^{\dag}$ & 77.25 & 78.30 & 91.17 & 87.85 & 80.18 & 83.00 & 72.87 & 81.52 \\
${\rm{Skip - though}}{{\rm{t}}^\dag }$ & 76.50 & 80.10 & 93.60 & 87.10 & 82.00 & 92.20 & 73.00 & 83.50 \\
\midrule[1pt]
${\rm{Avg}}{\rm{.\,  {BERT}\,  embedding}}{{\rm{s}}^\dag }$ & 78.66 & 86.25 & 94.37 & 88.66 & 84.40 & \textbf{92.80} & 69.54 & 84.94 \\ 
${\rm{BERT - }}\left[ {{\rm{CLS}}} \right]{\rm{embedding}}{{\rm{s}}^\dag }$ & 78.68 & 84.85 & 94.21 & 88.23 & 84.13 & 91.40 & 71.13 & 84.66 \\
${{\rm{IS - BER}}{{\rm{T}}_{{\rm{base}}}}^{\dag}}$ & 81.09 & \textbf{87.18} & 94.96 & 88.75 & 85.96 & 88.64 & 74.24 & 85.83 \\ 
${\rm{SCD - BER}}{{\rm{T}}_{{\rm{base}}}}^{\heartsuit}$ & 73.21 & 85.80 & \textbf{99.56} & 88.67 & 85.59 & 89.80 & 75.71 & 85.52\\
${{\rm{SimCSE - BER}}{{\rm{T}}_{{\rm{base}}}}}$ & 81.47 & 86.86 & 94.79 & 89.25 & 86.27 & 89.40 & 72.81 & 85.84 \\
${\rm{\texttt{InforMin-CL} - BER}}{{\rm{T}}_{{\rm{base}}}}$ & 80.99 & 85.72 & 94.63 & \textbf{89.47} & 85.67 & 88.20 & 73.97 & 85.52 \\
w/ MLM & \textbf{82.87} & 87.05 & 95.22 & 88.43 & \textbf{87.15} & 92.20 & \textbf{75.77} & \textbf{86.96}\\
\midrule[1pt]
${\rm{SimCSE - RoBERT}}{{\rm{a}}_{{\rm{base}}}}$ & 81.26 & 87.36 & 93.58 & 87.56 & 86.93 & 84.80 & 75.01 & 85.21 \\
${\rm{SCD - RoBERT}}{{\rm{a}}_{{\rm{base}}}}^{\heartsuit}$ & 82.17 & 87.76 & 93.67 & 85.69 & 88.19 & 83.40 & 76.23 & 85.30\\
${\rm{\texttt{InforMin-CL} - RoBERT}}{{\rm{a}}_{{\rm{base}}}}$ & 82.22 & 88.08 & 93.57 & \textbf{87.75} & 87.59 & 86.60 & 76.99 & 86.11 \\
w/ MLM & \textbf{83.49} & \textbf{88.69} & \textbf{94.79} & 86.81 & \textbf{88.30} & \textbf{89.40} & \textbf{77.57} & \textbf{87.01}\\
\midrule[1pt]
${\rm{SimCSE - RoBERT}}{{\rm{a}}_{{\rm{large}}}}{\mkern 1mu}$ & 80.85 & 85.99 & 93.08 & 87.65 & 86.33 & 89.00 & 72.46 & 85.05 \\
${\rm{\texttt{InforMin-CL} - RoBERT}}{{\rm{a}}_{{\rm{large}}}}$ & \textbf{82.50} & \textbf{88.32} & \textbf{93.81} & \textbf{89.38} & \textbf{87.64} & \textbf{90.80} & \textbf{74.49} & \textbf{86.71} \\
\bottomrule[2pt]
\end{tabular}}
\end{table*}

\section{Experiment}\label{section: experiment}
\subsection{Evaluation Setup}
We conduct our experiments on seven standard supervised tasks and also seven unsupervised tasks. We use the SentEval Toolkit \cite{conneau2018senteval} for evaluation. Following \cite{reimers2019sentence,Gao:Embeddings}, we take unsupervised tasks as the main comparison of the sentence embedding approaches and supervised results for reference. 

\textbf{Unsupervised Tasks} 
We evaluate representations on seven semantic textual similarity (STS) tasks: STS 2012-2016 \cite{agirre2012semeval,agirre2013sem,agirre2014semeval,agirre2015semeval,agirre2016semeval}, STS Benchmark \cite{cer2017semeval}, and SICK-Relatedness \cite{marelli2014sick} and compute the cosine similarity between sentence embeddings. All the unsupervised experiments are fully unsupervised, which means no STS training datasets are used and all embeddings are fixed once they are trained. For the sake of comparability, we follow the evaluation protocol of \cite{Gao:Embeddings}, employing Spearman's rank correlation and aggregation on all topic subsets.

\textbf{Supervised Tasks} 
We evaluate representations on seven supervised tasks: MR \cite{pang2005seeing}, CR \cite{hu2004mining}, SUBJ \cite{pang2004sentimental}, MPQA \cite{wiebe2005annotating}, SST-2 \cite{socher2013recursive}, TREC \cite{voorhees2000building} and MRPC \cite{dolan2005automatically}. A logistic regression classifier is trained on the top of (frozen) sentence embeddings produced by different methods. We follow default configurations from SentEval and use accuracy as the metric.

\textbf{Training Details} 
We start from pre-trained BERT \cite{devlin2019bert} (uncased) or RoBERTa \cite{liu2019roberta} (cased). Similar to \cite{Gao:Embeddings}, we train our \texttt{InforMin-CL} in an unsupervised fashion on $10^6$ randomly sampled sentences from English Wikipedia. During training, we add an MLP layer on the top of the [CLS] representation as sentence embeddings and directly take the [CLS] representation as sentence embeddings at testing time. A masked language modeling (MLM) objective \cite{devlin2019bert} is added as an optional auxiliary loss to the Eq. \ref{Eq loss}: ${\mathcal{L}} + \beta *{\mathcal{L}_{MLM}}$ ($\beta$ is a hyper-parameter).
For all results, we use the following hyper-parameters: epoch: 1, temperature $\tau$: 0.05, optimizer: Adam \cite{DBLP:journals/corr/KingmaB14}). We carry out grid-search of batch size $ \in $ $\left\{ {64,128,256} \right\}$ and learning rate $ \in $ $\left\{ {1e - 5,3e - 5,5e - 5} \right\}$ on STS-B development sets. During the training process, we save the checkpoint with the highest score on the STS-B development set to find the best hyperparameters. We adopt the hyperparameter settings listed in Table \ref{batch sizes, learning rates and lambda}. For all results, we use a PC with a GeForce RTX 3090 GPU (CUDA 11, PyTorch 1.7.1).

\subsection{Main Results}

\textbf{Baselines} We compare \texttt{InforMin-CL} to previous typical sentence embedding methods, which include averaging GloVe embeddings \cite{pennington2014glove}, Skip-thought \cite{kiros2015skip} and average BERT or RoBERTa embeddings. We also compare to post-processing methods and methods using a contrastive objective.  Post-processing methods include BERT-flow \cite{conf/emnlp/LiZHWYL20} and BERT-whitening \cite{su2021whitening}. Methods using a contrastive objective include IS-BERT \cite{zhang2020unsupervised}, DeCLUTR \cite{giorgi-etal-2021-declutr}, CT \cite{carlsson2020semantic}, SimCSE \cite{Gao:Embeddings}, and SCD \cite{klein2022scd}. IS-BERT maximizes the agreement between global and local features. 
DeCLUTR takes different spans from the same document as positive pairs. 
CT aligns embeddings of the same sentence from two different encoders. 
SimCSE takes the embedding of the same input with different dropouts as positive pairs. 
SCD leverages the self-contrast of augmented samples obtained by dropout.

\textbf{Performance on Unsupervised Tasks}
Table \ref{Unsupervised downstream task results} shows the evaluation results on seven STS tasks. \texttt{InforMin-CL} achieves comparable or better results than previous state-of-the-art baselines. 
For ${\rm{BER}}{{\rm{T}}_{{\rm{base}}}}$ based models, our method outperforms the best approach with a large margin (\textbf{$ + 1.67\% $}) on average.
For RoBERTa based model, \texttt{InforMin-CL} obtains comparable results (less than 0.4 points in average). 
All these indicate that our model can improve the performance of downstream tasks by forgetting the irrelevant information in pre-training phase.

\textbf{Performance on Supervised Tasks}
Table \ref{Supervised downstream task results} shows the evaluation results on seven supervised tasks. Results indicate that \texttt{InforMin-CL} performs on par or better than all baselines. 
Our method achieves better results on most of and even all tasks using RoBERTa, with an average gain $0.81\%$ on ${\rm{RoBERT}}{{\rm{a}}_{{\rm{base}}}}$ and $1.66\%$ on ${\rm{RoBERT}}{{\rm{a}}_{{\rm{large}}}}$ respectively. 
With an MLM task added, further gains on average results are observed for BERT and RoBERTa. It raises the average scores of \texttt{InforMin-CL} from $ 85.52\%$ to $ 86.96\%$ for ${\rm{BER}}{{\rm{T}}_{{\rm{base}}}}$ and from $ 86.11\%$ to $ 87.01\%$ for ${\rm{RoBERT}}{{\rm{a}}_{{\rm{base}}}}$. 
Particularly, \texttt{InforMin-CL} w/ MLM obtains the outperforms all the baselines in average.

Considering results of both supervised and unsupervised tasks, we present the following findings: 1) BERT-based \texttt{InforMin-CL} performs better on unsupervised tasks; 
2) RoBERTa-based \texttt{InforMin-CL} achieves better results on supervised tasks. 
The difference in performance using BERT and RoBERTa is mainly caused by the difference in pre-training corpus. 
BERT is trained over 16 GB text (BooksCorpus \cite{zhu2015aligning} and English Wikipedia) while RoBERTa is trained over totally 160 GB of uncompressed text (BooksCorpus \cite{zhu2015aligning}, English Wikipedia, CC-NEWS \cite{nagel2016cc}, OPENWEBTEXT \cite{gokaslan2019openwebtext}, and STORIES \cite{trinh2018simple}).
Thus, the diverse large-scale high quality datasets enhance the RoBERTa to learn the important and useful information with limited parameters. 
\texttt{InforMin-CL}, which improves performance by discarding redundant information, struggles to represent its effects in this setting due to less noise information. 
This causes the unsupervised results of \texttt{InforMin-CL} are similar to competitors for RoBERTa-based models.

\subsection{Ablation Study}
\paragraph{\textbf{Influence of $\lambda$}} We investigate how different reconstruction objectives with $\lambda$ from 0.04 to 4 affect our model's performance. We report the average performance of unsupervised tasks and supervised tasks in this experiment. The results are obtained using ${\rm{BER}}{{\rm{T}}_{{\rm{base}}}}$. Results demonstrate that \texttt{InforMin-CL} constantly works well over this wide range of $\lambda$. As shown in Table \ref{hyper-parameter}, with increasing $\lambda$, the performance of both unsupervised and supervised tasks rises first and falls later.

\paragraph{\textbf{Influence of $\beta$}} We introduce one more optional variant which adds a masked language modeling (MLM) objective to the Eq. \ref{Eq loss}: ${\mathcal{L}} + \beta *{\mathcal{L}_{MLM}}$ ($\beta$ is a hyper-parameter). We analyze how different $\beta$ influence the performance on unsupervised and supervised tasks. As we show in Table \ref{hyper-parameter beta}, we find that adding MLM objectives with different $\beta$ consistently helps improve performance on supervised tasks but brings a significant drop in STS tasks.

\paragraph{\textbf{Influence of Batch Sizes}}
To explore the impact of batch sizes, we report the average performance of downstream tasks with batch sizes (N in Eq. \ref{Eq 5}) from 64 to 256. In this experiment, only batch size changes while all other hyper-parameters keep unchanged. We use ${\rm{BER}}{{\rm{T}}_{{\rm{base}}}}$ to evaluate on the test set of unsupervised and supervised tasks. As we show in Table \ref{ablation batch size}, we find that \texttt{InforMin-CL} is not sensitive to batch size, similar to SimCSE, mainly caused by the good set of initial parameters.


\begin{table}[h]
    \centering
    \caption{Ablation studies of different hyper-parameters $\lambda$.
    The results are based on the test sets using ${\rm{BER}}{{\rm{T}}_{{\rm{base}}}}$.}
    \label{hyper-parameter}
    \begin{tabular}{p{20px}p{75px}<{\centering}p{75px}<{\centering}}
    \toprule[2pt]
    $\lambda$ & \textbf{Avg. Sup} & \textbf{Avg. Unsup} \\
    \midrule[1pt]
    0.04 & 85.20 & 76.09 \\
    0.4 & \textbf{85.52} & \textbf{77.30} \\
    4 & 85.03 & 77.18\\
    \bottomrule[2pt]
    \end{tabular}
\end{table}

\begin{figure*}[!h]
    \centering
    \subfigure[SimCSE]{
    \label{simcse}
    \begin{minipage}[t]{0.3\linewidth}
    \includegraphics[width=1.65in]{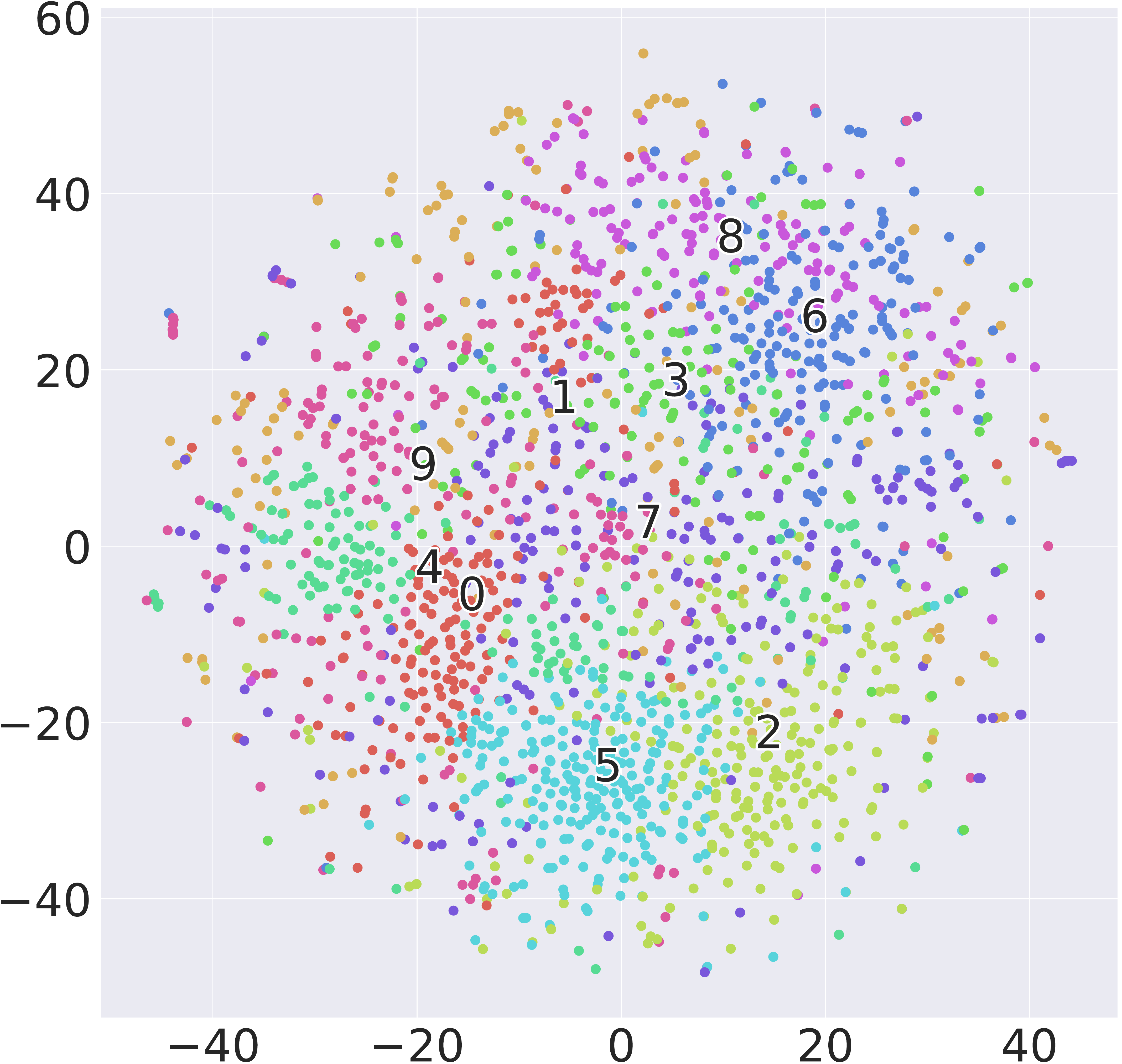}
    \end{minipage}
    }
    \subfigure[SCD]{
    \label{scd}
    \begin{minipage}[t]{0.3\linewidth}
    \includegraphics[width=1.65in]{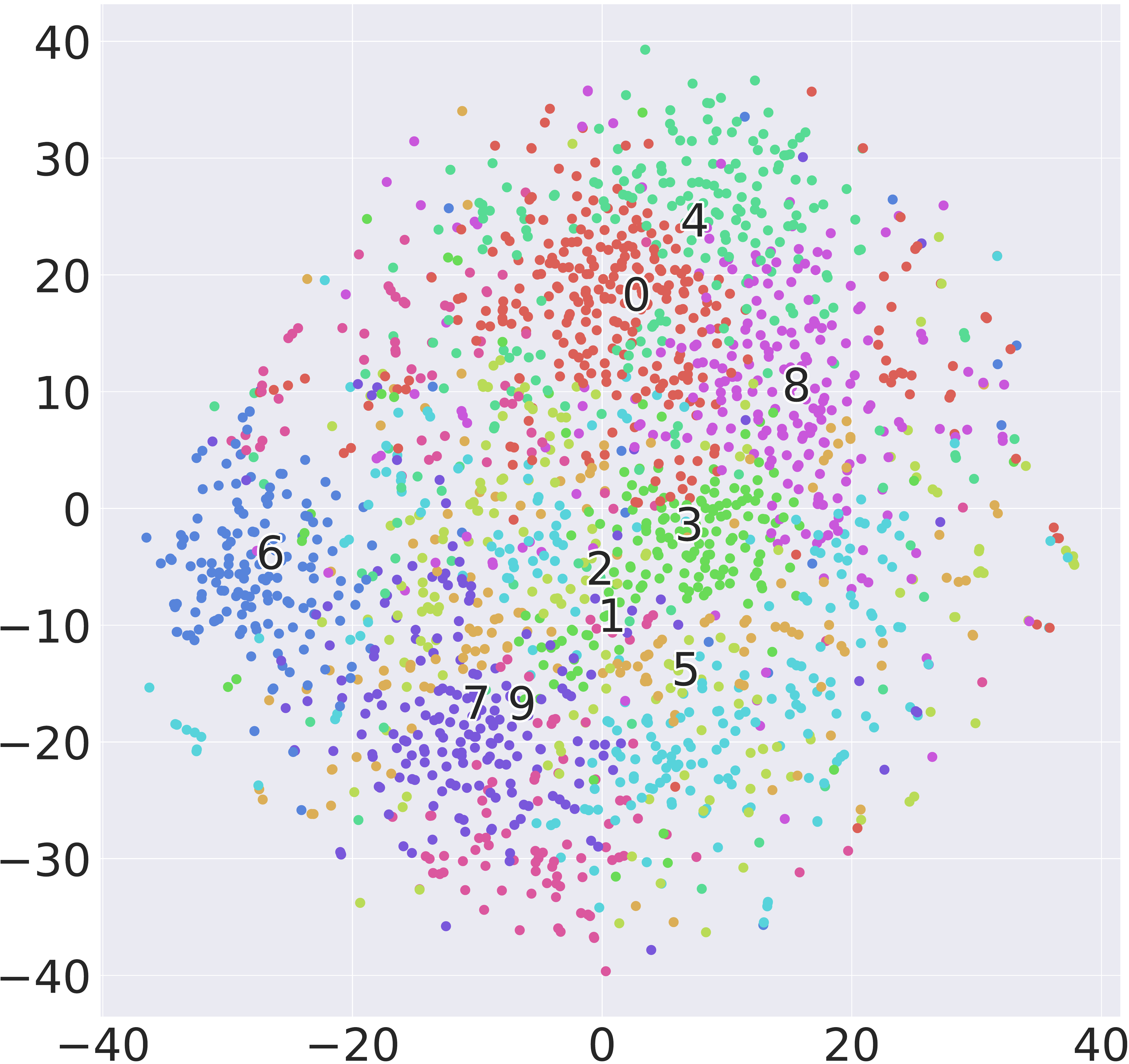}
    \end{minipage}
    }
    \subfigure[\texttt{InforMin-CL}]{
    \label{informin-cl}
    \begin{minipage}[t]{0.3\linewidth}
    \includegraphics[width=1.65in]{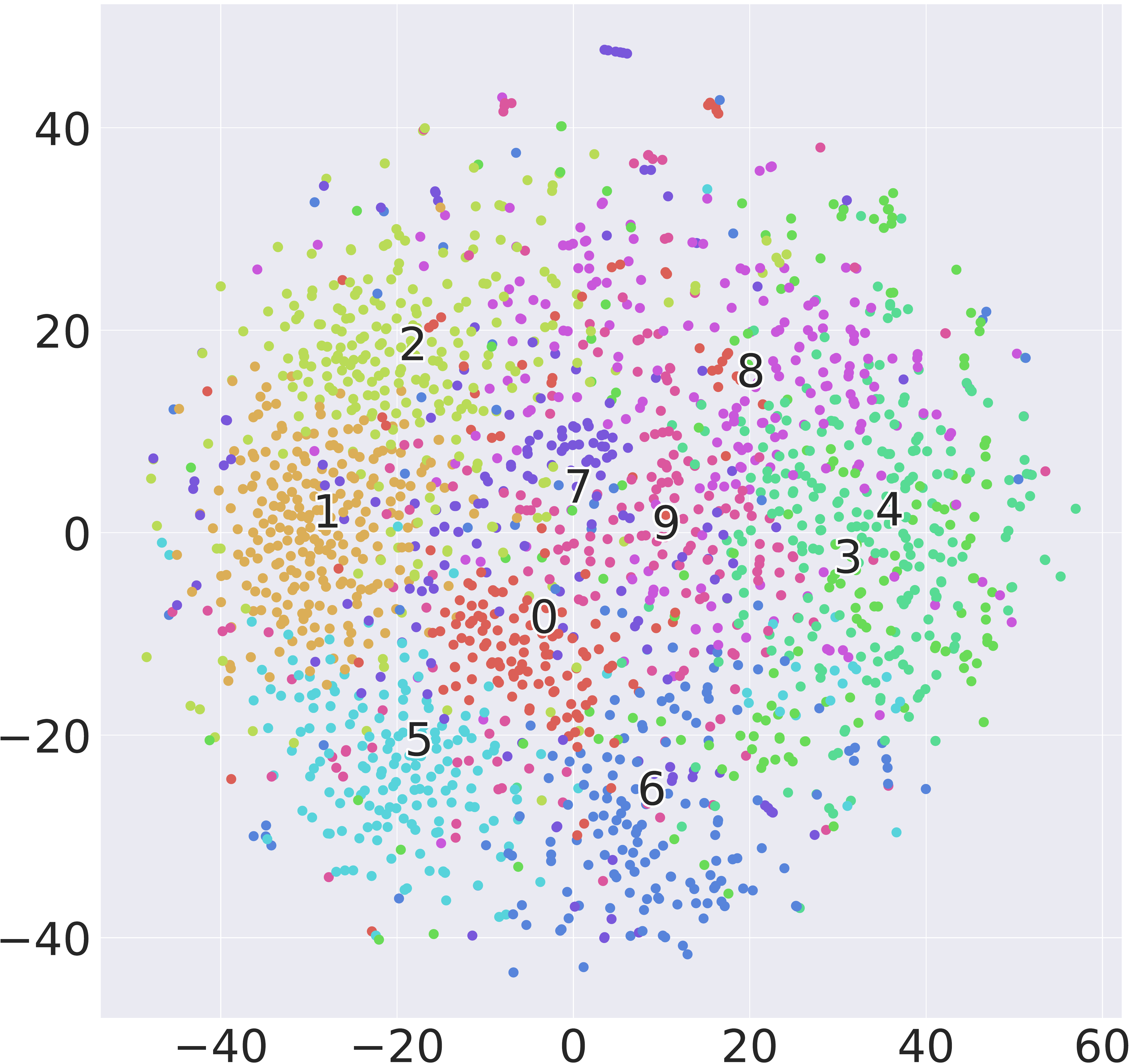}
    \end{minipage}
    }
    \caption{The t-SNE of sentence representations learned with SimCSE, SCD and \texttt{InforMin-CL} using ${\rm{BER}}{{\rm{T}}_{{\rm{base}}}}$. The points are embeddings of sentences sampled from the IMDB dataset without fine-tuning.}
    \label{Fig2-5}
\end{figure*}

\begin{figure}[!ht]
    \centering
    \includegraphics[width=0.5\textwidth]{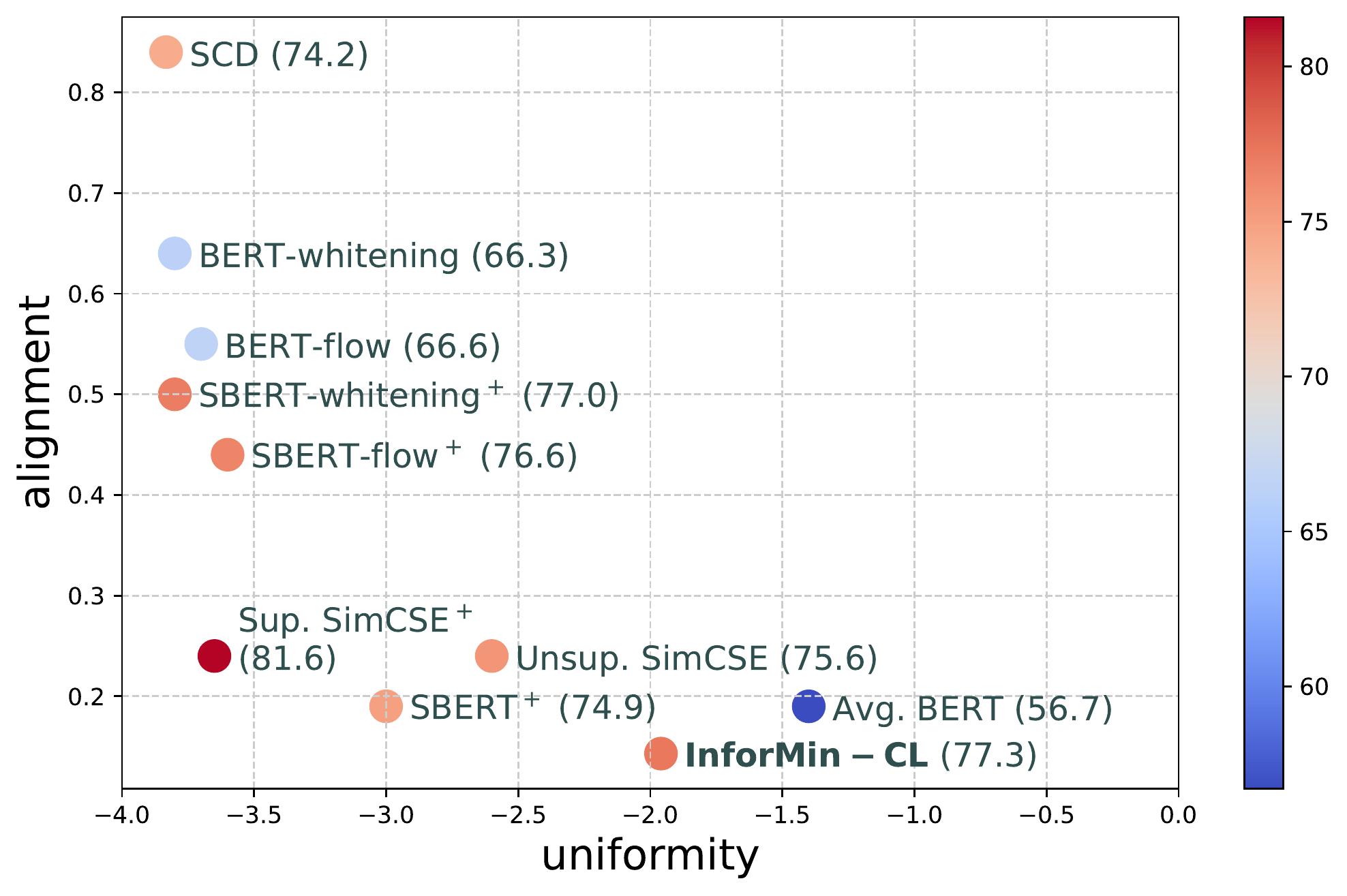}
    \caption{Quantitative analysis of embeddings - \textit{alignment} vs. \textit{uniformity} (the smaller, the better). The plot of models is based on ${\rm{BER}}{{\rm{T}}_{{\rm{base}}}}$. Points represent average STS performance with Spearman's correlation color coded ($ + $ corresponds to supervised methods).}
    \label{Fig4}
\end{figure}

\begin{table}[h]
    \centering
    \caption{Ablation studies of the MLM objective based on the test sets using ${\rm{BER}}{{\rm{T}}_{{\rm{base}}}}$.}
    \label{hyper-parameter beta}
    \begin{tabular}{lcc}
    \toprule[2pt]
    \textbf{Model} & \textbf{Avg. Sup} & \textbf{Avg. Unsup} \\
    \midrule[1pt]
    w/o MLM & 85.52 & \textbf{77.30}\\
    w/ MLM & {} & {}\\
    $\,\,\beta$ = 0.01 & 86.46 & \textbf{63.59}\\
    $\,\,\beta$ = 0.1 (ours) & 86.96 & 63.25\\
    $\,\,\beta$ = 1.0 & \textbf{87.04} & 60.85\\
    \bottomrule[2pt]
    \end{tabular}
\end{table}

\begin{table}[h]
    \centering
    \caption{Ablation studies of different batch sizes. The results are based on test sets using ${\rm{BER}}{{\rm{T}}_{{\rm{base}}}}$.}
    \label{ablation batch size}
    \begin{tabular}{p{75px}p{30px}<{\centering}p{30px}<{\centering}p{30px}<{\centering}}
    \toprule[2pt]
    \textbf{Batch size} & \textbf{64} & \textbf{128} & \textbf{256} \\
    \midrule[1pt]
    Avg. Sup & 85.38 & 85.52 & \textbf{85.77} \\
    Avg. Unsup & 76.64 & \textbf{77.30} & 76.14\\
    \bottomrule[2pt]
    \end{tabular}
\end{table}

\vspace{-1mm}
\subsection{Uniformity and Alignment}
We further conduct analysis to understand the inner workings of \texttt{InforMin-CL}.
\paragraph{\textbf{Qualitative Analysis}}
As shown in \cite{wang2020understanding}, the asymptotics of the contrastive learning objective (\ref{Eq 5}) can be expressed by the following equation when the number of negative instances approaches infinity:

\begin{equation}
\begin{array}{l}
\begin{split}
{\mathcal{L}_{C}} =& \max\, \Big [\frac{1}{N}\sum\limits_{i = 1}^N {\mathbb{E}\big[ {{{sim\left( {z_i^1,z_i^2} \right)} \mathord{\left/
 {\vphantom {{sim\left( {z_i^1,z_i^2} \right)} \tau }} \right.
 \kern-\nulldelimiterspace} \tau }} \big]} \\
 &- \frac{1}{N}\sum\limits_{i = 1}^N {\mathbb{E}\, \big[ {\log \frac{1}{N}\sum\limits_{k = 1}^N {{e^{{{sim\left( {z_i^1,z_k^2} \right)} \mathord{\left/
 {\vphantom {{sim\left( {z_i^1,z_k^2} \right)} \tau }} \right.
 \kern-\nulldelimiterspace} \tau }}}} } \big]} \Big]
\end{split}
\end{array}
\label{equation transform}
\end{equation}

The first term in square brackets in Eq. \ref{equation transform} improves alignment of the space. The alignment performs better when the similarity score rises. While optimizing the reconstruction objective, $z^1$ and $z^2$ are pulled closer, which means that the similarity score of $z^1$ and $z^2$ becomes higher. In other words, \texttt{InforMin-CL} effectively improves alignment of pre-trained embeddings while keeping a good uniformity, which is the key to the success of \texttt{InforMin-CL}. We also follow \cite{wang2020understanding} to use uniformity and alignment to measure the quality of representation space for \texttt{InforMin-CL} and other models. Figure \ref{Fig4} shows uniformity and alignment of different sentence embedding models along with their STS averaged results. \texttt{InforMin-CL} achieves the best in terms of \textit{alignment} (0.143), which can be related to the strong effect of the reconstruction objective. In terms of \textit{uniformity}, \texttt{InforMin-CL} is slightly inferior to unsupervised SimCSE. This is also reflected in the final results in the t-SNE plots. 

\paragraph{Quantitative Analysis}
The t-SNE \cite{reif2019visualizing} plot in Figure \ref{Fig2-5} demonstrates the advantages of \texttt{InforMin-CL}. We sample 2000 sentences from IMDB \cite{maas2011learning} dataset and generate the embeddings of sentences using SimCSE, SCD and InforMin-CL. We use K-Means \cite{jain1988algorithms} clustering to group similar sentence embeddings and form 10 clusters. Results indicate that similar sentence pairs (marked by same colors) generated by \texttt{InforMin-CL} are more aligned.

\section{Limitations}
Although our method outperforms baselines on both unsupervised and supervised tasks in most cases, there are still at least two limitations. First, we simply sample negative instances from other sentences in the mini-batch, which may lead to false negatives. Punishing false negatives during training by assigning lower weight for negatives with higher similarity may be a solution. 
Second, although redundant information is discarded, what redundant information forgets and remains is unknown.
It would be interesting to explore this problem by integrating interpretation methods. 

\section{Conclusion}
In this work, we propose \texttt{InforMin-CL}, an effective contrastive learning approach, which improves state-of-the-art sentence embedding performance on downstream tasks. \texttt{InforMin-CL} discards redundant information stored in positive instances by encouraging one positive instance to reconstruct the other positive instance in the same pair. We test \texttt{InforMin-CL} on seven supervised and seven unsupervised tasks. Experimental results indicate our method outperforms all previous competitors.

\section*{Acknowledgements}
We thank the anonymous reviewers for their valuable comments. We also thank Shanghai Science and Technology Innovation Action Plan Project (21511104500) for funding this research.

\bibliography{anthology,custom}

\end{document}